\documentclass{article}
\usepackage{ijcai25}

\usepackage{times}                
\usepackage[utf8]{inputenc}       
\usepackage{textcomp}             
\usepackage{relsize}              
\usepackage{stfloats}             

\usepackage{amsmath}              
\usepackage{amsthm}               
\usepackage{amssymb}              
\usepackage{algorithm}            
\usepackage{algorithmic}          

\usepackage{booktabs}             
\usepackage{multirow}             
\usepackage{array}                
\usepackage{caption}              
\usepackage{graphicx}             
\usepackage{rotating}             
\usepackage{tikz}                 
\usetikzlibrary{calc}            
\usepackage{tikz}

\usepackage[hidelinks]{hyperref}  
\usepackage{url}                  

\usepackage[switch]{lineno}       

\usepackage{soul}                 

\usepackage{subcaption}           

\title{GPI-Net: Gestalt-Guided Parallel Interaction Network via Orthogonal Geometric Consistency for Robust Point Cloud Registration}

\author{
Weikang Gu$^1$\and
Mingyue Han$^1$\and
Li Xue$^1$\And
Heng Dong$^2$\And
Changcai Yang$^1$\And
Riqing Chen$^1$\And
Lifang Wei$^1$
\\
\affiliations
$^1$College of Computer and Information Science, Fujian Agriculture and Forestry University\\
$^2$School of Computing and Information Science, Fuzhou Institute of Technology\\
\emails
gwk154137@163.com,
mingyue.han@163.sufe.edu.cn,
395391275@qq.com,
changcaiyang@gmail.com,
\{52311049004, riqing.chen, weilifang0028\}@fafu.edu.cn
}

\definecolor{MyBlue}{HTML}{488B87} 

\begin{document}

\maketitle

\begin{abstract}
    The accurate identification of high-quality correspondences is a prerequisite task in feature-based point cloud registration. However, it is extremely challenging to handle the fusion of local and global features due to feature redundancy and complex spatial relationships. Given that Gestalt principles provide key advantages in analyzing local and global relationships, we propose a novel Gestalt-guided Parallel Interaction Network via orthogonal geometric consistency (GPI-Net) in this paper. It utilizes Gestalt principles to facilitate complementary communication between local and global information. Specifically, we introduce an orthogonal integration strategy to optimally reduce redundant information and generate a more compact global structure for high-quality correspondences. To capture geometric features in correspondences, we leverage a Gestalt Feature Attention (GFA) block through a hybrid utilization of self-attention and cross-attention mechanisms.  Furthermore, to facilitate the integration of local detail information into the global structure, we design an innovative Dual-path Multi-Granularity parallel interaction aggregation (DMG) block to promote information exchange across different granularities. Extensive experiments on various challenging tasks demonstrate the superior performance of our proposed GPI-Net in comparison to existing methods. The code will be released at https://github.com/gwk429/GPI-Net.
\end{abstract}

\section{Introduction}

\begin{figure}[ht]
\centering
\captionsetup{justification=justified}
\captionsetup{labelformat=default, labelsep=colon}
\includegraphics[width=\linewidth]{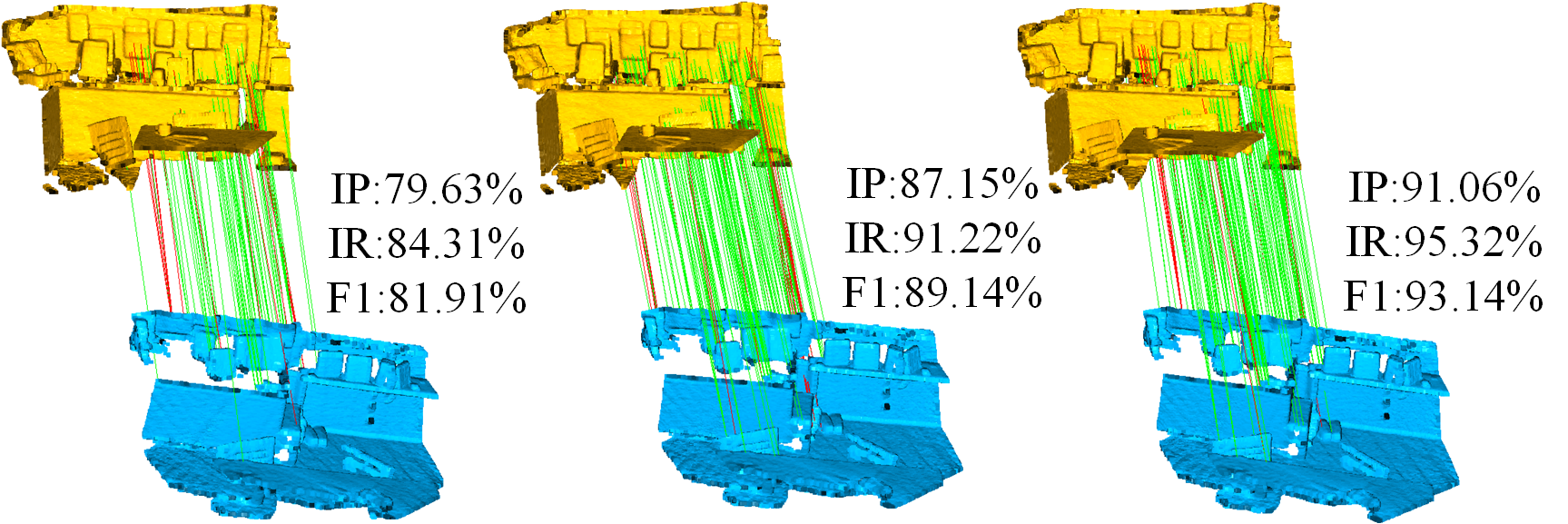} \\

\begin{tabular}{>{\centering\arraybackslash}m{0.3\linewidth} 
                >{\centering\arraybackslash}m{0.3\linewidth} 
                >{\centering\arraybackslash}m{0.3\linewidth}}  
    (a) PointDSC & (b) PG-Net & (c) GPI-Net \\
\end{tabular}
\caption{Visualization results of outlier removal. The green and red lines highlight inliers and outliers, respectively. On account of  the power of our proposed GPI-Net to optimally aggregate local details and global information across multiple granularities, it demonstrates superior performance in identifying more inliers.}
\label{fig:example1}  
\end{figure}

The point cloud is now the predominant data format for the representation of the 3D world. In addition, a variety of point cloud processing algorithms have been designed for diverse applications, such as simultaneous localization and mapping~ \cite{xue_stereo_2022}, autonomous driving ~\cite{okuyama_autonomous_2018}, and robotics \cite{zhou_loop_2022}. Sensors like LiDAR ~\cite{zhao_lif-seg_2024} and Microsoft Kinect ~\cite{yang_applying_2016} are capable of generating point clouds directly. However, due to their limited line of sight, these sensors must be repositioned to obtain data from multiple viewpoints. Point cloud registration has been developed to ensure the precise alignment of point clouds collected from different perspectives. Feature-based point cloud registration typically starts by generating initial correspondences utilizing feature descriptors, such as Fast Point Feature Histograms (FPFH) ~\cite{rusu_fast_2009} and Fully Convolutional Geometric Feature (FCGF) ~\cite{choy_fully_2019}. Nevertheless, owing to the lack of clarity in the feature descriptors and the ambiguity of local features, the initial correspondences frequently contain a substantial number of erroneous correspondences, giving rise to inaccurate alignment. Consequently, developing a robust and high-accuracy outlier removal network is crucial for increasing the inlier ratio and achieving accurate point cloud registration.

In recent decades, researchers have proposed various excellent outlier removal methods, including RANSAC~ \cite{fischler_random_1981}, Spectral Matching (SM) \cite{leordeanu_spectral_2005}, Fast Global Registration (FGR) ~\cite{leibe_fast_2016}, CG-SAC ~\cite{quan_compatibility-guided_2020}, and TEASER \cite{yang_teaser_2021}. Although these methods are effective in simple scenarios, they struggle to converge in complex scenes with high outlier ratios. It brings about inaccurate alignment under elevated outlier ratios. Recently, deep learning-based outlier removal methods have improved the accuracy and robustness of point cloud registration with stronger feature representation. Specifically, DGR \cite{choy_deep_2020} adopts a feature embedding network to better extract global context information. 3DRegNet ~\cite{pais_3dregnet_2020} introduces classification and registration modules to recognize inliers. However, these methods rely solely on multilayer perceptrons and simple sparse convolution to capture the contextual information of correspondences, overlooking the importance of spatial consistency among inliers in 3D space. To overcome these challenges, PointDSC ~\cite{bai2021pointdsc} incorporates spatial consistency information to improve the accuracy of point cloud registration. PG-Net ~\cite{wang_pg-net_2023} designs a grouped dense fusion attention module to collect rich contextual information and uses inlier probabilities to guide the classification of initial correspondences. Although the aforementioned methods perform well, they are insufficient to grasp the relationship between local details and global information. Figure \ref{fig:example1}(a) and (b) show that these methods left more outliers, which hinders effective complementarity and interaction among correspondences and may leave excessive redundant information. In summary, existing methods inevitably leave redundant information when exploring the relationship between local and global information, making it difficult to fully aggregate features and effectively capture more contextual information. To tackle these challenges, this paper proposes a novel Gestalt-guided Parallel Interaction Network via orthogonal geometric consistency (GPI-Net) designed for more robust and accurate outlier removal.

The Gestalt principles ~\cite{xu_unsupervised_2018} are visual perception principles that emphasize the superiority of whole perception over local perception, where the meaning of the whole is greater than the sum of the individual parts. 
In computer vision, object recognition is not solely based on the features of individual details but rather involves perceiving their overall structure and context.
Leveraging Gestalt principles, GPI-Net not only extracts comprehensive and effective geometric features to support the acquisition of high-quality correspondences but also facilitates the complementarity and communication between local and global information, which reduces the generation of redundant information.
Specifically, we introduce an Orthogonal Integration (OI) strategy in GPI-Net. The OI utilizes the spatial relationships of feature vectors to better aggregate local and global information while reducing redundancy during feature fusion. 
Then, we design a Gestalt Feature Attention (GFA) block based on the closure in Gestalt principles. 
The GFA accurately extracts geometric features of correspondences and effectively supplements missing contextual information by alternately applying self-attention ~\cite{gu_salient_2021} and cross-attention ~\cite{kuang_towards_2023}, leveraging the relationships between surrounding features.
Following the core ideas of the Gestalt principles, we present a novel Dual-path Multi-Granularity parallel interaction aggregation (DMG) block, which integrates contextual information from multiple granularities to better acquire local details and efficiently fuse them into global structural information. The GPI-Net boosts the differentiation between inliers and outliers to enhance understanding the internal dependencies of correspondences. Comprehensive experiments on outlier removal and pose estimation tasks reveal that our proposed framework exceeds existing methods.

We summarize the main contributions as follows.

\begin{itemize}
    \item We propose a novel GPI-Net to address the feature redundancy and complex spatial relationships, optimally guiding the classification of correspondences and alleviating the impact of outliers on point cloud registration.
    \item We incorporate Gestalt principles to guide GPI-Net in accurately identifying inliers on outlier removal task by emphasizing the superiority of whole perception over local perception.
    \item We design the OI and GFA to reduce redundant information by the spatial geometric features of correspondences, respectively. The DMG integrates spatial information across granularities via bidirectional hierarchical interaction, further enhancing the efficient fusion of local and global information.  
\end{itemize}

\section{Related Works}

In this section, we briefly introduce some point cloud registration and outlier removal methods.

\subsection{Point Cloud Registration}

Iterative Closest Point (ICP) ~\cite{besl_method_1992} is a classical traditional point cloud registration algorithm. The core idea of ICP is to iteratively optimize the objective function employing least squares to estimate the optimal rigid transformation. ICP is a more suitable choice when the point cloud already has good initial alignment or when there are fewer outliers in the scene. Nevertheless, traditional methods struggle to balance registration accuracy and computational cost. As a result, deep learning techniques have emerged to address these challenges.
PointNet ~\cite{charles_pointnet_2017} proposes a groundbreaking work that utilizes neural networks to extract features from point cloud. The core idea of PointNet is to employ symmetric functions to process unordered point cloud. However, PointNet has some limitations, particularly in extracting local geometric information. It neglects the local structure of point clouds and fails to fully utilize local information.
On the basis of the limitations of PointNet in local feature learning, PointNet++ ~\cite{qi_pointnet_2017} is presented to improve the network's ability to perceive local structures and extract multi-scale features. By introducing a hierarchical feature extraction structure, PointNet++ effectively overcomes the shortcomings of PointNet in local feature learning and multi-scale feature representation.

\subsection{Outlier Removal}

RANSAC ~\cite{fischler_random_1981} achieves its goal by repeatedly selecting a random subset of the data. The selected subset is assumed to consist of inliers and the least-squares method is utilized for model fitting. Numerous RANSAC variants have introduced new sampling strategies and local optimization techniques to improve robustness. For instance, Locally Optimized RANSAC (LO-RANSAC) ~\cite{goos_locally_2003} enhances robustness in the presence of local outliers by processing local data regions.
However, RANSAC and its variants typically rely on random sampling and model evaluation, bringing about slow convergence and low accuracy.
DHVR ~\cite{lee_deep_2021} employs a decoupled multi-model fitting architecture based on deep Hough voting, separating and sequentially estimating rotation and translation. PG-Net ~\cite{wang_pg-net_2023} designs a grouped dense fusion attention module to gather rich contextual information and an iterative structure to guide the classification of initial correspondences via inlier probabilities. However, these networks are insufficient to grasp the spatial relationship between local details and global information. Meanwhile, gathering spatial information at multiple granularities remains highly challenging. In this paper, our proposed GPI-Net uses a dual-path parallel interaction strategy with Gestalt principles, achieving exchange between local and global information across different granularities to improve pose estimation performance.

\section{Method}

\begin{figure*}
    \centering
    \captionsetup{justification=justified}
    \includegraphics[width=\textwidth]{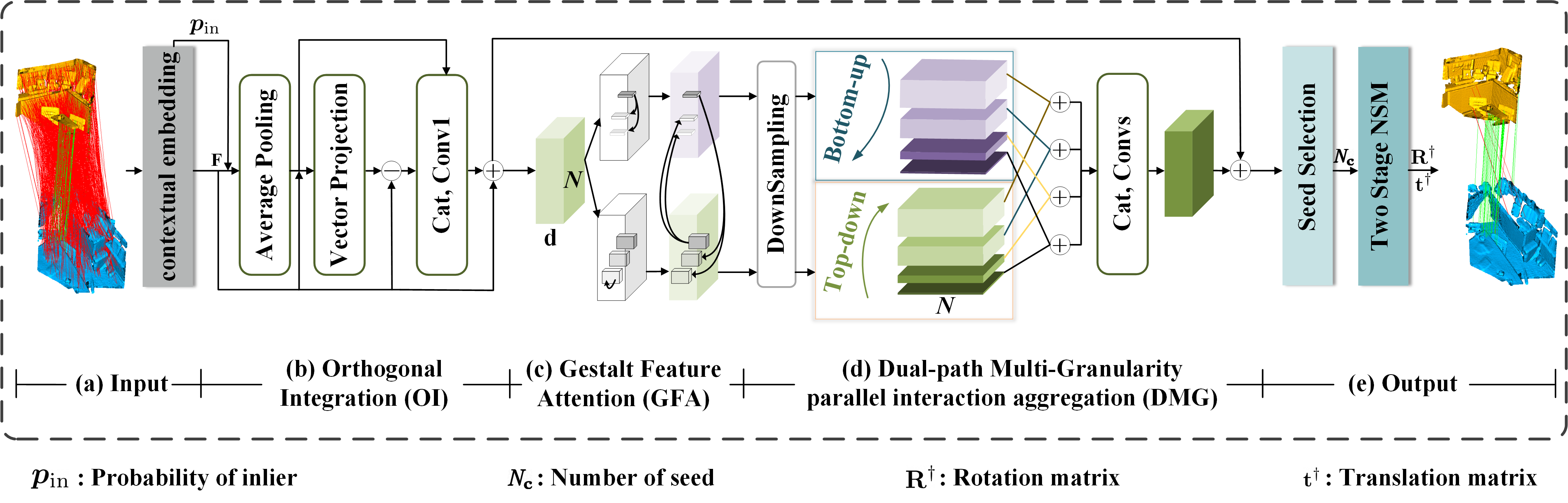} 
    \caption{The framework of our proposed GPI-Net. GPI-Net takes the initial correspondences as input and generates $N_c$ transformation matrices. It selects the optimal transformation matrix based on the number of inliers, achieving precise point cloud alignment. The OI, GFA, and DMG form the key components of GPI-Net.}
    \label{fig:example2}  
\end{figure*}

In the following parts, we first introduce the problem formulation, then provide an overview of our proposed network, followed by a detailed explanation of each block.

\subsection{Problem Formulation}

Point cloud registration aims to align two partially overlapping point clouds, $P_s = \{ p_k^s \in \mathbb{R}^3 \mid k = 1, 2, \ldots, N \}$  and $P_t = \{ p_j^t \in \mathbb{R}^3 \mid j = 1, 2, \ldots, M \}$, into the same coordinate space by finding the transformation matrix between them. Firstly, we extract keypoint features from the point clouds via local descriptors (such as FPFH and FCGF) and generate the initial correspondences set $C$ from these features through nearest-neighbor search.
\begin{equation}
C = \{c_1, c_2, \ldots, c_N\} \in \mathbb{R}^{N \times 6}, \quad c_i = (p_i^s, p_i^t) \in C,
\end{equation}
where $c_i$ is the $i$th initial correspondence. $p_i^s \in P_s$ and $p_i^t \in P_t$
represent the spatial coordinates of the keypoints in the source and target point clouds. $N$ is the number of initial correspondences.

Next, the initial correspondences are mapped to the probability that each correspondence is an inlier in the feature space, obtaining high-quality correspondences. At last, the high-quality correspondences are utilized to estimate the rigid transformation matrix, achieving point cloud registration.

Specifically, GPI-Net learns the probability of each correspondence being an inlier in high-dimensional feature space. High-confidence correspondences are identified as seeds, and each seed obtains a consensus set in the metric space. Following the Two-Stage NSM ~\cite{wang_pg-net_2023}, each consensus set produces a transformation matrix $\mathbf{R}', \mathbf{t}'$. Through GPI-Net, we obtain $N_c$ rigid transformation set $\mathbf{Rt}_{set} = \{ (\mathbf{R}_1', \mathbf{t}_1'); (\mathbf{R}_2', \mathbf{t}_2'); \ldots; (\mathbf{R}_{N_c}', \mathbf{t}_{N_c}') \}$, and the best transformation matrix $\mathbf{R}^{\dagger}, \mathbf{t}^{\dagger}$. The selection strategy is based on the number of correct correspondences to evaluate the quality of each transformation matrix. The transformation result of $\mathbf{R}^{\dagger}$ and $\mathbf{t}^{\dagger}$ is computed from the given set:
\begin{equation}
\mathbf{R}^{\dagger}, \mathbf{t}^{\dagger} = \underset{(\mathbf{R}', \mathbf{t}')\in\mathbf{Rt}_{set}}{\arg\max} \sum_{i=1}^{N} \left[ \| \mathbf{R}' p_i^s + \mathbf{t}' - p_i^t \| < \delta \right],
\end{equation}
where [$\cdot$] is the Iverson bracket, $\delta$ represents a predefined threshold for the inlier, and \( \left\| \cdot \right\| \) denotes the Euclidean norm. 

\subsection{Overview}

Our GPI-Net is shown in Figure \ref{fig:example2}. Initially, we input
$N$ pairs of initial correspondences into our proposed GPI-Net. The
contextual embedding module \cite{wang_pg-net_2023} maps the initial correspondences into a high-dimensional feature space and preliminarily extracts contextual information through grouped feature fusion. Next, the OI effectively filters out redundant information generated during feature aggregation by orthogonal integration. Then, the GFA leverages self-attention and cross-attention to further gather the geometric features of correspondences in high-dimensional space, enriching the contextual information. On this basis, the DMG collects features and facilitates information exchange across different granularities through a dual-path parallel interaction strategy, promoting the complementarity and interaction between local and global information. Subsequently, the seed selection module identifies high-quality correspondences. Finally, the Two-Stage NSM module is used to estimate the optimal point cloud transformation matrix.

\subsection{Orthogonal Integration (OI)}


To develop robust interactions between local details and global structures, we provide a feature map $F = \{ f_p \}_{p=1}^N$, where \textit{N} represents the number of correspondences. It is necessary to gain global features, which are commonly obtained via global average pooling. However, this approach is unsuitable for the outlier removal task, on the ground that global average pooling assigns equal importance to each correspondence, while outliers have a detrimental effect on the network.
To tackle this challenge, we adopt a weighted global average pooling to gather global features as shown in Figure \ref{fig:example2}(b). Specifically, we first map the feature map \textit{F} to the inlier probability $\boldsymbol{w}$ of each correspondence by convolution. The probability value measures the importance of each correspondence within the network, guiding it to embed stronger global contextual information. The process is formalized as follows:
\begin{equation}
F_g = \xi(F, \boldsymbol{w}) = \sum_{p=1}^N \frac{w_p}{\sum_{q=1}^N w_q} f_p,
\end{equation}
here, $\xi$ represents the weighted global average pooling operation. Then, we pass $F_g$ through a bottleneck layer consisting of two 1D convolutional layers to further generate a more robust global feature $F_g^\tau$. Following this, we consider how to more optimally fuse local and global features while minimizing redundant information. Traditional methods simply employ operations like concatenation or element-wise addition, yet these methods frequently introduce considerable redundancy. Therefore, we apply orthogonal integration to achieve more compact feature aggregation. Specifically, we compute the projection of \textit{F} in the direction of $F_g^\tau$, denoted as $F_{\text{projection}}$. Mathematically, the projection is as follows:
\begin{equation}
F_{\text{projection}} = \frac{F \cdot F_g^\tau}{\|F_g^\tau\|^2} \cdot F_g^\tau.
\end{equation}

Owing to the difference in orthogonal components between the feature $\mathit{F}$ and the projection vector $F_{\text{projection}}$, we further obtain the more crucial local detail feature map $\mathit{f}$ within $\mathit{F}$, guided by the feature information in $F_{\text{projection}}$. The process is written as follows:
\begin{equation}
f = F - F_{\text{projection}}.
\end{equation}

By this way, we gather a more crucial local information feature map from \textit{F}, generating a more robust local feature representation \textit{f}. The \textit{f} not only contains more detailed feature information among correspondences, but also significantly reduces the interference of redundant information. Subsequently, we perform a concatenation operation between the \textit{f} and $F_g^\tau$, followed by the convolutional layer to produce a more compact global feature representation $F_o$. The process is formalized as follows:
\begin{equation}
F_o = \textit{Conv1} \big( \text{Cat}(f, F_g^T) \big) + F,
\label{eq:example6}
\end{equation}
where \textit{Conv1(·)} indicates a convolution kernel with a size of 1×1. Through Equation (\ref{eq:example6}), the network precisely captures the critical features of the task while maintaining optimal information representation capabilities through efficient integration of both local and global information.

\subsection{Gestalt Feature Attention (GFA)}

To comprehensively collect the geometric features among correspondences, we leverage the core idea of the Gestalt principles: holistic perception is superior to local perception. As illustrated in Figure \ref{fig:example2}(c),  we integrate self-attention and cross-attention mechanisms to identify similarities both within and between feature maps for capturing geometric information.
The self-attention acquires long-range dependencies and global information by modeling relationships among correspondences. The cross-attention dynamically links connections between different feature maps, efficiently blending information from multiple feature maps. This ensures that information obtained from different feature maps is effectively intertwined and complementary.
The GFA learns geometric information and complex relationships from local dependencies to global associations across different levels by stacking self-attention and cross-attention. 

To tackle these complex variations, we combine the weights of different sizes, aiming to fully integrate the intrinsic local information into global structure. 
Specifically, the feature $F_o$ is passed through a convolutional layer to generate the query (\( \mathbf{Q} \)), key (\( \mathbf{K} \)), and value (\( \mathbf{V} \)). We compute self-attention across various correspondences and across multiple feature maps, using $W_1$ of size \( \mathbb{R}^{N \times N} \) for correspondence-level attention and $W_2$ of size \( \mathbb{R}^{\mathrm{d} \times \mathrm{d}} \) for feature-map-level attention. Here, $N$ represents the number of correspondences and d denotes the number of feature channels.
\begin{align}
f_{at_1} = \textit{PWConv}\left(\text{softmax}\left(\mathbf{Q}\cdot\mathbf{K}^T\right) \cdot \mathbf{V}\right) + F_o, \\[1em]
f_{at_2} = \textit{PWConv}\left(\text{softmax}\left(\mathbf{Q}^T\cdot\mathbf{K}\right) \cdot \mathbf{V}^T\right)^T + F_o,
\end{align}
where $PWConv(\cdot)$ denotes a point-wise convolution layer, $f_{at1}$ represents the feature information obtained through the weight $W_1$, which captures the interactions and dependencies among correspondences on a global granularity. The global information facilitates the network to understand the similarities and differences between correspondences from a holistic perspective. In contrast, $f_{at2}$  focuses on collecting the internal structural information of features by calculating the similarity across each feature dimension, which emphasizes the dependencies between feature dimensions.

Self-attention struggles to fully explore the interactions and complementarities between different feature spaces. Based on the closure in Gestalt principles, missing information is supplemented by leveraging relationships between surrounding features. To boost dynamic connections and ensure complementarity between $f_{at1}$and $f_{at2}$, we adopt a cross-attention mechanism to alleviate the limitations of representations based on a single feature space. The cross-attention calculates the similarity between the \( \mathbf{Q} \) and \( \mathbf{K} \) from different feature maps, dynamically adjusting the weighted representation of the \( \mathbf{V} \). 
Specifically, we utilize $f_{at1}$ as the Q, $f_{at2}$ as the K and V to generate robust feature $f_{at1}'$, which incorporates contextual information from $f_{at2}$. Then, we reverse the roles to generate robust feature $f_{at2}'$ that absorbs information from $f_{at1}$. The process enhances the information flow between different feature maps.

By combining self-attention and cross-attention, the network obtains geometric feature information and dependencies from a local to a global perspective. 


\subsection{Dual-Path Multi-Granularity Parallel Interaction Aggregation (DMG)}

Extracting features at multiple granularities is highly challenging due to the complexity of the features and the difficulty in effectively balancing local and global features of the data. Grounded in the Gestalt principles that the whole is greater than the sum of its parts, we design a DMG block. The DMG adopts a dual-path parallel interaction strategy, achieving information exchange and integration between local and global features across different granularities.

Figure \ref{fig:example2}(d) shows that the features $f_{at1}'$ and 
$f_{at2}'$ from the GFA are downsampled into 
$T$ granularities via average pooling. For $f_{at1}'$, the multi-granularity feature set is represented as $F_T$ = $ \{f_0, \cdots, f_T\} $, where $f_t \in \mathbb{R}^{N \times \mathrm{d} / 2^t}$, and $t \in \{0, \cdots, T\}$. $N$ and d represent the number of correspondences and feature channels, respectively. Similarly, the $f_{at2}'$ undergoes the same operation to generate a multi-granularity feature set $G_T$ = $ \{g_0, \cdots, g_T\} $.

Gestalt principles suggest that relationships between local elements shape the perception of the whole. DMG integrates local and global information across granularities via bidirectional hierarchical interaction to produce a richer representation. Therefore, it is necessary to fuse features of different granularities to avoid the potential bias and to facilitate interaction between local and global information. Specifically, we apply dual-path mixing operations to different feature maps for promoting information exchange across multiple granularities. The process is defined by:
\begin{equation}
F_{\text{Ges}} = \textit{Convs}\big(\text{Cat}(\{B\_Mix(F_t) + T\_Mix(G_t)\}_{t=0}^T)\big),
\end{equation}
where the $Convs(\cdot)$ operation includes an InstanceNorm normalization layer, a BatchNorm layer, a ReLU activation function, a pointwise convolution layer, and a Shuffle channel mixing layer. The pointwise convolution layer takes input features of size $\mathbb{R}^{N \times 15\mathrm{d}/8}$ and produces output of size $\mathbb{R}^{N \times \mathrm{d}}$. $B\_Mix(\cdot)$ and $T\_Mix(\cdot)$ represent the coarse-to-fine mixing layer and fine-to-coarse mixing layer, respectively. 

$\boldsymbol{B\_Mix}(\cdot)$ \textbf{layer}: Coarse-grained features provide a global and macro-level perspective, smoothing out local noise and facilitating the acquisition of global structural information by the network. Building on this, we adopt a bottom-up mixing strategy, starting with global features and gradually incorporating fine-grained local information, which guides the network to more accurately understand local details. Technically, for the multi-granularity feature set $F_T$ = $ \{f_0, \cdots, f_T\} $, we utilize a residual structure to apply the $B\_Mix(\cdot)$ layer to the $t$-th granularity, which achieves bottom-up mixing to facilitate effective information exchange between correspondences. The operation is formalized as follows:
\begin{equation}
F_t = F_t + \textit{Mix\_Bottom\_up}(F_{t-1}),
\end{equation}
where $t \in \{1, \cdots, T\}$ and $\textit{Mix\_Bottom\_up}(\cdot)$ consists of an InstanceNorm layer, a BatchNorm layer, a ReLU activation function, and a pointwise convolution layer. The pointwise convolution layer takes input and output features of size $\mathbb{R}^{N \times \mathrm{d} / 2^{t-1}}$ and $\mathbb{R}^{N \times \mathrm{d} / 2^{t}}$, respectively.

$\boldsymbol{T\_Mix}(\cdot)$ \textbf{layer}: The network incorporates broader context and global information to maintain sensitivity to details after capturing fine-grained geometric features and structures. This fine-to-coarse strategy accurately captures local details first and gradually integrates global information, enhancing overall representation. Technically, for the multi-granularity features $G_T$ = $ \{g_0, \cdots, g_T\} $, we also utilize a residual structure to employ the $T\_Mix(\cdot)$ layer to the $t$-th granularity, forming a top-down mixing strategy. The operation is by:
\begin{equation}
G_t = G_t + \textit{Mix\_Top\_down}(G_{t+1}),
\end{equation}
where $t \in \{1, \cdots, T\}$ and $\textit{Mix\_Top\_down}(\cdot)$ consists of an InstanceNorm layer, a BatchNorm layer, a ReLU activation function, and a pointwise convolution layer. 
The pointwise convolution layer takes input and output features of size $\mathbb{R}^{N \times \mathrm{d} / 2^{t+1}}$ and $\mathbb{R}^{N \times \mathrm{d} / 2^{t}}$, respectively.

By introducing the DMG, the coarse-grained global perspective effectively smooths out noise in the fine-grained features. At the same time, the block extracts subtle geometric features at the fine-grained level and merges them with coarse-grained features to gather the global structure. This dual-path parallel interaction strategy, combining coarse-to-fine and fine-to-coarse approaches, has the power to achieve more precise global alignment.

\section{Experiments}

In this section, we first introduce the datasets and experimental parameters employed to verify the effectiveness of GPI-Net. Next, we evaluate GPI-Net in indoor and outdoor scenarios via FPFH and FCGF descriptors, respectively. Finally, we conduct an ablation study regarding our GPI-Net.

\renewcommand{\arraystretch}{1.5}
\newcommand{\thickhline}{\noalign{\hrule height 1.2pt}}
\newcolumntype{C}[1]{>{\centering\arraybackslash}m{#1}}

\begin{table*}[htbp]
\centering
\resizebox{\textwidth}{!}{ 
\begin{tabular}{c c C{0.9cm} C{0.9cm} C{0.9cm} C{0.9cm} C{0.9cm} C{0.9cm} C{0.9cm} | C{0.9cm} C{0.9cm} C{0.9cm} C{0.9cm} C{0.9cm} C{0.9cm} C{0.9cm}}
\thickhline
\multirow{2}{*}{~} & \multirow{2}{*}{Method} & \multicolumn{7}{c}{\textbf{FPFH}} & \multicolumn{7}{c}{\textbf{FCGF}} \\ \cmidrule(lr){3-16}
 & & RR(\%$\uparrow$) & RE($^\circ\downarrow$) & TE(cm$\downarrow$) & IP(\%↑) & IR(\%↑) & F1(\%↑) & Time(s) & RR(\%$\uparrow$) & RE($^\circ\downarrow$) & TE(cm$\downarrow$) & IP(\%↑) & IR(\%↑) & F1(\%↑) & Time(s) \\
\midrule
\multirow{5}{*}{\rotatebox{90}{Traditional}} 
& SM           & 55.88 & 2.94 & 8.15 & 47.96 & 70.69 & 50.70 & \textbf{0.03} & 86.57 & 2.29 & 7.07 & 81.44 & 38.36 & 48.21 & \textbf{0.03} \\ 
& FGR          & 40.91 & 4.96 & 10.25 & 6.84 & 38.90 & 11.23 & 0.14 & 78.93 & 2.90 & 8.41 & 25.63 & 53.90 & 33.58 & 0.17 \\
& TEASER       & 75.48 & 2.48 & 7.31 & 73.01 & 62.63 & 66.93 & 0.04 & 85.77 & 2.73 & 8.66 & \textbf{82.43} & 68.08 & 73.96 & 0.11 \\
& RANSAC-1M    & 64.20 & 4.05 & 11.35 & 63.96 & 57.90 & 60.13 & 0.39 & 88.42 & 3.05 & 9.42 & 77.96 & 79.86 & 78.55 & 0.45\\
& RANSAC-2M    & 65.25 & 4.07 & 11.56 & 64.41 & 58.37 & 60.51 & 3.87 & 90.88 & 2.71 & 8.31 & 78.52 & 83.52 & 80.68 & 4.13\\
\cmidrule(lr){2-16}
\multirow{5}{*}{\rotatebox{90}{Learning}} 
& 3DRegNet     & 26.31 & 3.75 & 9.60 & 28.21 & 8.90 & 11.63 & 0.05 & 77.76 & 2.74 & 8.13 & 67.34 & 56.28 & 58.33 & 0.05\\
& DGR          & 32.84 & 2.45 & 7.53 & 29.51 & 16.78 & 21.35 & 2.49 & 88.85 & 2.28 & 7.02 & 68.51 & 79.92 & 73.15 & 1.36 \\
& DHVR         & 67.10 & 2.78 & 7.84 & 60.19 & 64.90 & 62.11 & 0.46 & 90.93 & 2.25 & 7.08 & 78.35 & 78.15 & 78.94 & 0.46 \\
& PointDSC     & 78.56 & \textbf{2.06} & \textbf{6.48} & 68.92 & 71.96 & 70.17 & 0.09 & 92.73 & 2.04 & 6.49 & 78.83 & 86.07 & 82.02 & 0.09 \\
& PG-Net       & 81.96 & 2.07 & 6.62 & 72.45 & 77.72 & 74.82 & 0.18 & 92.98 & 2.04 & 6.47 & 79.12 & 86.03 & 82.16 & 0.18 \\
\cmidrule(lr){2-16}
& OURS         & \textbf{83.25} & 2.08 & 6.57 & \textbf{73.06} & \textbf{78.47} & \textbf{75.57} & 0.19 & \textbf{93.15} & \textbf{2.02} & \textbf{6.47} & 79.35 & \textbf{86.28} & \textbf{82.47} & 0.19 \\
\thickhline
\end{tabular}
}
\caption{Quantitative results on the 3DMatch dataset for outlier rejection and pose estimation.}
\label{tab:example1}
\end{table*}

\subsection{Datasets and Experimental Setup}

Regarding indoor scenarios, we follow the same evaluation protocol as 3DMatch \cite{zeng_3dmatch_2017} to prepare the training and testing datasets. The testing set consists of 1623 pairs of point clouds from 8 indoor scenes. For each pair, we downsample the point clouds employing a 5cm voxel grid. For outdoor scenarios, we utilize the KITTI dataset \cite{geiger_vision_2013} to test the model's effectiveness. The testing set comprises 555 pairs of point clouds from 10 outdoor scenes, with each pair downsampled using a 30cm voxel grid.

We assess the effectiveness of GPI-Net through two tasks: outlier rejection and pose estimation. Regarding outlier rejection, we apply inlier precision (IP\%), inlier recall (IR\%), and F1 score (F1\%) as evaluation metrics to measure the accuracy of inliers identification while minimizing false positives. For pose estimation, we adopt registration recall (RR\%), rotation error (RE), and translation error (TE). Registration is considered successful when both RE and TE are below the specified thresholds: RE $<$ 15° and TE $<$ 30cm for indoor scenarios, and RE $<$ 5° and TE $<$ 60cm for outdoor scenarios. RE and TE are reported only on successfully registered point clouds.

\subsection{Evaluation of Indoor Scenes}

We first present the experimental results on the 3DMatch dataset to test the performance of our proposed GPI-Net in indoor scenarios. Table \ref{tab:example1} presents the comparative results on the 3DMatch dataset.

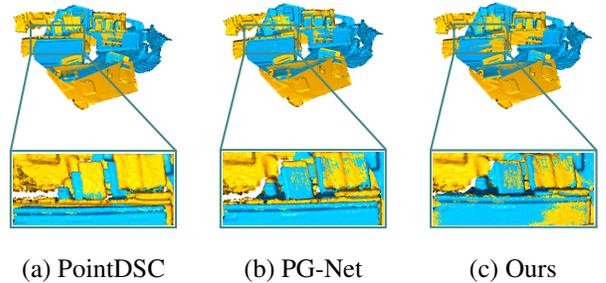
\begin{figure}[ht]
\centering
\captionsetup{justification=justified}
\captionsetup{labelformat=default, labelsep=colon}
\begin{tabular}{ccc}
    
    \begin{tikzpicture}
        \node[anchor=south west] (image) at (0,0) {\includegraphics[width=0.12\textwidth]{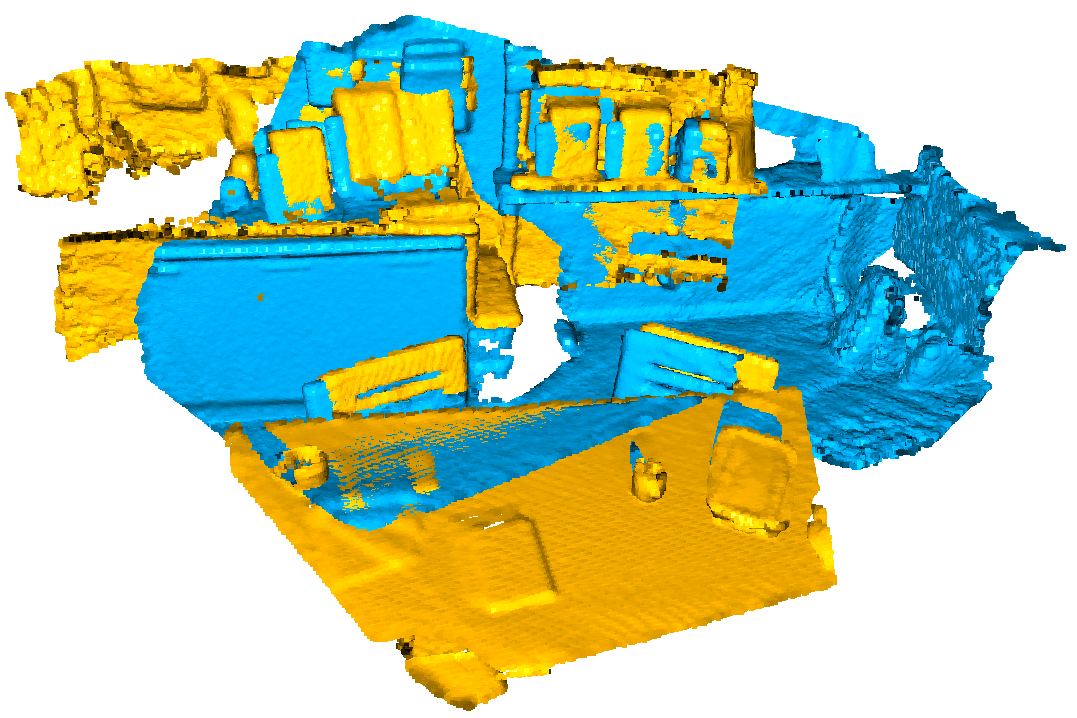}};
        \begin{scope}[x={(image.south east)},y={(image.north west)}]
            \draw[color=MyBlue,thick] (0.15,0.52) rectangle (0.45,0.75); 
            \draw[color=MyBlue,thick] (0.038,-0.25) rectangle (0.96,-0.86);  
            \node[anchor=north west] (zoomed) at (0,-0.2) {\includegraphics[width=0.12\textwidth]{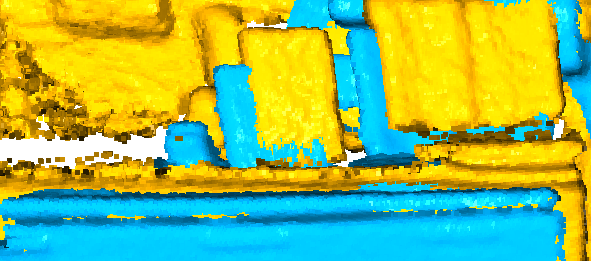}};
            \draw[thick, color=MyBlue] (0.15,0.52) -- (0.038,-0.25);
            \draw[thick, color=MyBlue] (0.45,0.52) -- (0.96,-0.25);
        \end{scope}
    \end{tikzpicture} &

    \begin{tikzpicture}
        \node[anchor=south west] (image) at (0,0) {\includegraphics[width=0.12\textwidth]{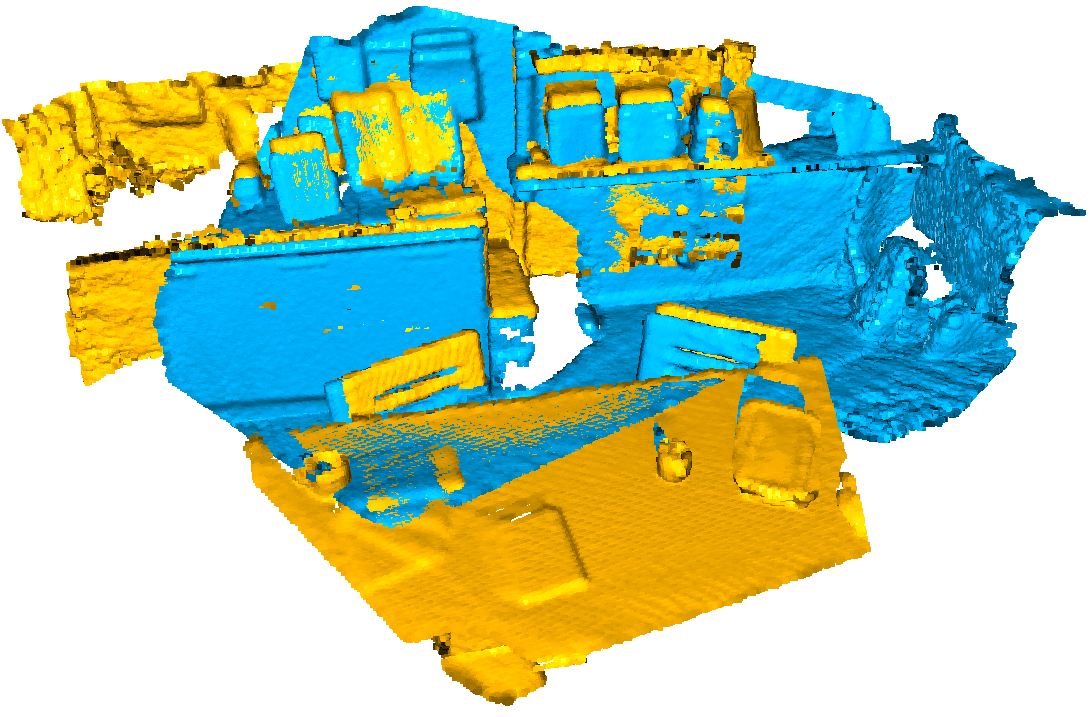}};
        \begin{scope}[x={(image.south east)},y={(image.north west)}]
            \draw[color=MyBlue,thick] (0.17,0.52) rectangle (0.47,0.75); 
            \draw[color=MyBlue,thick] (0.038,-0.255) rectangle (0.96,-0.88);  
            \node[anchor=north west] (zoomed) at (0,-0.2) {\includegraphics[width=0.12\textwidth]{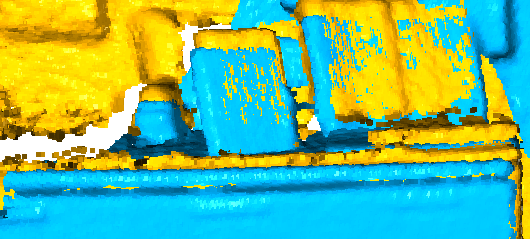}};
            \draw[thick, color=MyBlue] (0.17,0.52) -- (0.038,-0.255);
            \draw[thick, color=MyBlue] (0.47,0.52) -- (0.96,-0.25);
        \end{scope}
    \end{tikzpicture} &

    \begin{tikzpicture}
        \node[anchor=south west] (image) at (0,0) {\includegraphics[width=0.12\textwidth]{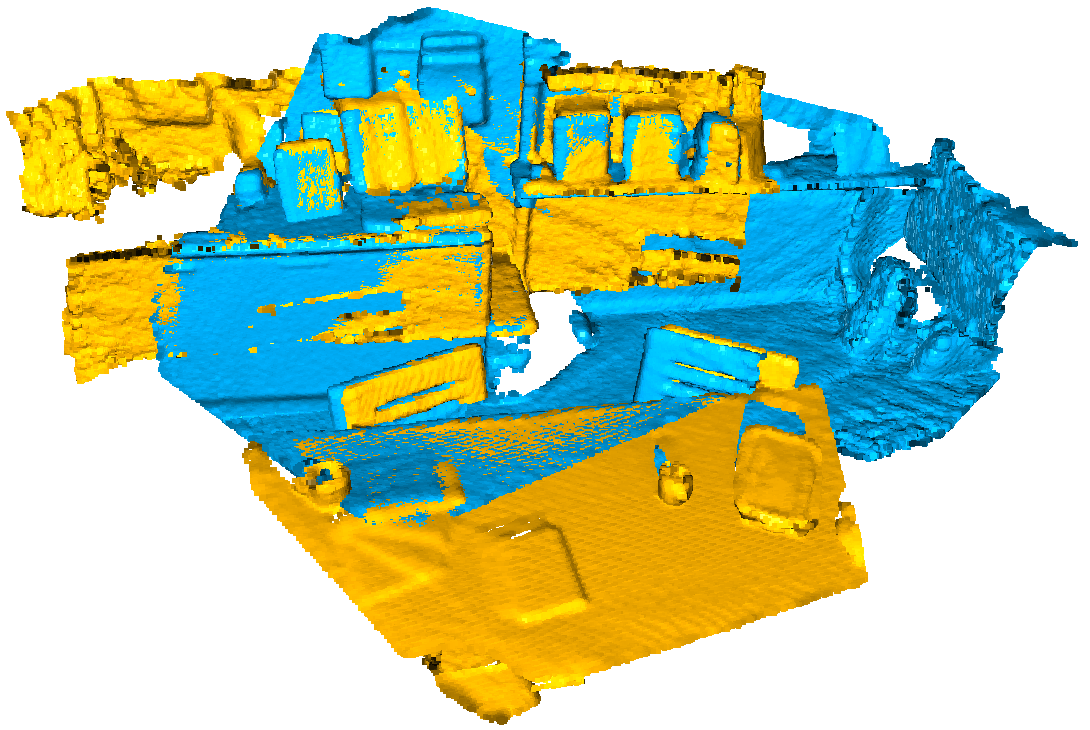}};
        \begin{scope}[x={(image.south east)},y={(image.north west)}]
            \draw[color=MyBlue,thick] (0.17,0.52) rectangle (0.47,0.75); 
            \draw[color=MyBlue,thick] (0.04,-0.25) rectangle (0.965,-0.86);  
            \node[anchor=north west] (zoomed) at (0,-0.2) {\includegraphics[width=0.12\textwidth]{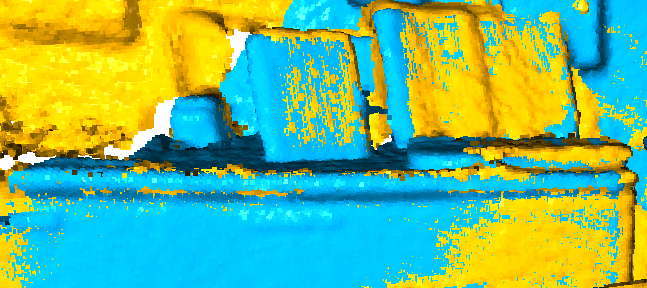}};
            \draw[thick, color=MyBlue] (0.17,0.52) -- (0.04,-0.25);
            \draw[thick, color=MyBlue] (0.47,0.52) -- (0.965,-0.25);
        \end{scope}
    \end{tikzpicture} \\
    
    (a) PointDSC &  (b) PG-Net & (c) Ours \\
\end{tabular}
\caption{Qualitative comparison on 3DMatch dataset.}
\label{fig:example3}  
\end{figure}

\paragraph{Combined with FPFH.}We first adopt FPFH descriptor to generate initial correspondences. From Table \ref{tab:example1}, we see that our method demonstrates satisfactory improvements in F1 and RR compared to other methods. Specifically, our method achieves an F1 approximately 6\% higher than PointDSC and 1\% higher than PG-Net, fully corroborating its effectiveness in the outlier rejection task. In terms of RR, our method surpasses PointDSC by nearly 5\% and PG-Net by approximately 1.5\%, further highlighting its superiority in the pose estimation task. This proves that GPI-Net, guided by Gestalt principles, optimally captures the complex relationships and internal dependencies between correspondences, leading to the acquisition of richer global structural information. As shown in Figure \ref{fig:example3}, we conduct a qualitative comparison with FPFH and our method reaches more accurate point cloud alignment relative to PointDSC and PG-Net.
In contrast to traditional point cloud registration, GPI-Net outperforms RANSAC-2M in RR by nearly 20\%, while maintaining low rotation and translation errors. Furthermore, we observe that 3DRegNet performs significantly worse than our approach in terms of RR with a gap of 56.94\%. This indicates that using a simple MLP as the feature extractor is significantly limited in capturing global information. These limitations lead to poor representation of inliers, bringing about inaccurate point cloud alignment. It confirms that GPI-Net excels in inlier identification as well as in achieving lower registration errors, indicating an improvement in overall alignment accuracy.

\paragraph{Combined with FCGF.} To comprehensively confirm the effectiveness of GPI-Net, we re-measure all methods exploiting FCGF descriptor. The results demonstrate that the FCGF descriptor consistently exhibits exceptional robustness and performance in different tasks. Our GPI-Net outperforms PG-Net in terms of F1 and RR, reaching 82.47\% and 93.15\%, respectively, which strongly affirms the superior performance of GPI-Net leveraging the FCGF descriptor.
Moreover, we observe that while TEASER achieves a slightly higher IP than GPI-Net, GPI-Net significantly surpasses TEASER in IR and F1. It indicates that TEASER has limited capability in identifying inliers and struggles to collect structural information of inliers over a broader scope. Such a limitation affects its accuracy and robustness in point cloud registration. In contrast, GPI-Net demonstrates greater stability and effectiveness in the outlier removal task.
Additionally, we find that employing the more robust FCGF descriptor consistently leads to better performance than the FPFH descriptor in both outlier rejection and pose estimation tasks. 
Table \ref{tab:example1} reports that although our network experiences a slight decline in RR when applying FPFH in place of FCGF, the reduction is only 9.9\%, whereas PG-Net shows a decrease of 11.02\%. This underscores the robustness and adaptability of GPI-Net to different initial correspondences generated by varying descriptors.

\begin{figure}
    \centering
    \captionsetup{justification=justified}
    \includegraphics[width=\columnwidth]{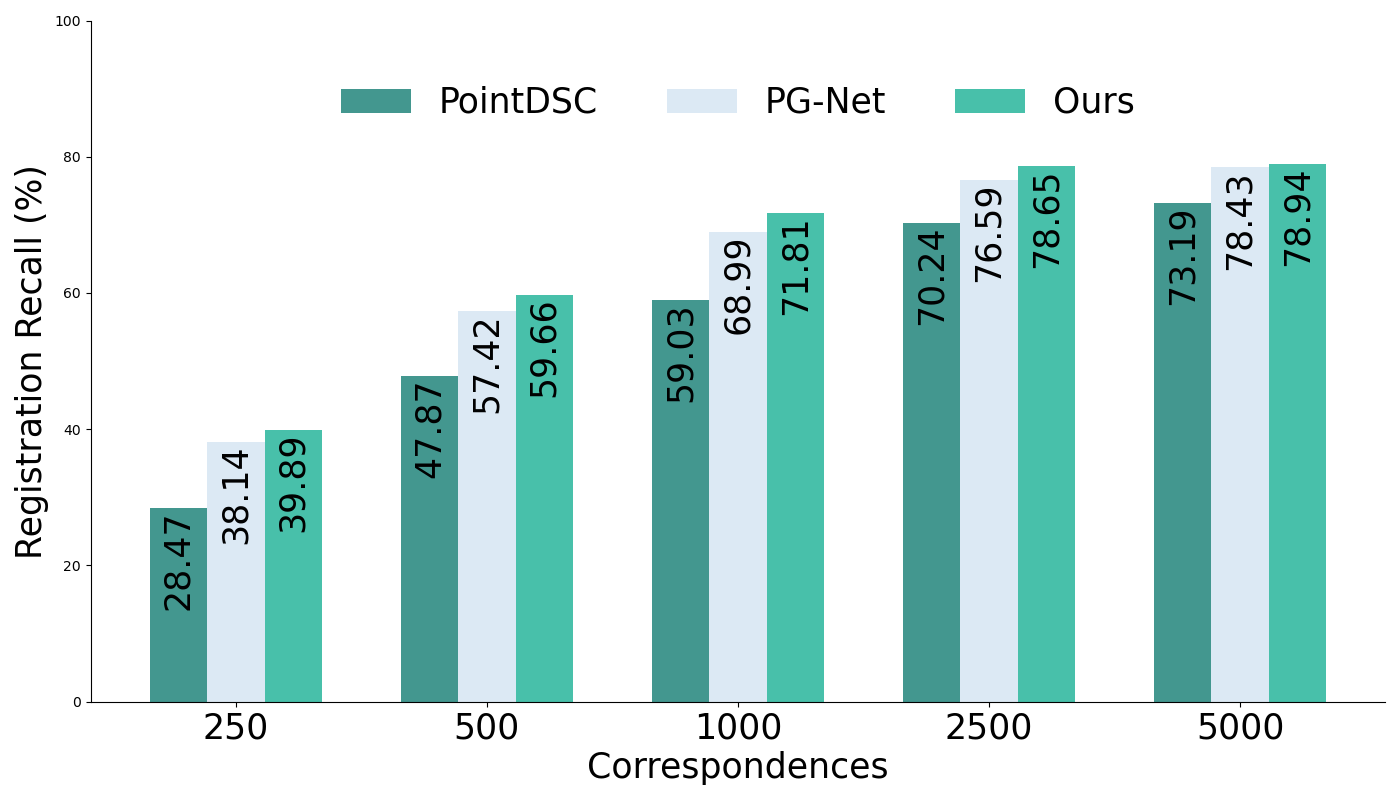} 
    \caption{Comparison of RR utilizing different numbers of correspondences with FPFH.}
    \label{fig:example4}  
\end{figure}

\paragraph{Robustness with fewer correspondences.} In the pose estimation task, the accuracy heavily depends on the number of correspondences. Therefore, to additionally verify the effectiveness and robustness of GPI-Net, we investigate the impact of the number of correspondences and compare it with two methods: PointDSC and PG-Net. Figure \ref{fig:example4} shows that GPI-Net exhibits an RR that exceeds PointDSC by more than 10\% at correspondence counts of 250, 500, and 1000. Similarly, GPI-Net surpasses PG-Net by over 2\% when the number of correspondences is 500, 1000, and 2500. Overall, GPI-Net consistently outperforms PointDSC and PG-Net regardless of the number of correspondences. Notably, GPI-Net demonstrates increasingly pronounced advantages as the number of correspondences decreases, underscoring its effectiveness in scenarios with fewer correspondences.

\subsection{Evaluation of Outdoor Scenes}

\renewcommand{\arraystretch}{2}

\begin{table}[tb]
    \centering
    \resizebox{\columnwidth}{!}{ 
    \Huge 
    \begin{tabular}{cccccccccc}
    \noalign{\hrule height 6pt} 
        \multirow{2}{*}{~} & \multirow{2}{*}{Method} & \multicolumn{4}{c}{\textbf{FPFH}} & \multicolumn{4}{c}{\textbf{FCGF}} \\ \cline{3-10}\cline{3-10}\cline{3-10}
        & & RR(\%↑) & RE(°↓) & TE(cm↓) & F1(\%↑) & RR(\%↑) & RE(°↓) & TE(cm↓) & F1(\%↑) \\ \cline{2-10}\cline{2-10}\cline{2-10}
         \multirow{3}{*}{\rotatebox{90}{Traditional}}  
        & FGR & 5.23 & 0.86 & 43.84 & — & 89.54 & 0.46 & 25.72 & — \\ 
        & RANSAC & 74.41 & 1.55 & 30.20 & 73.13 & 80.36 & 0.73 & 26.79 & 85.42 \\ 
        & CG-SAC & 74.23 & 0.73 & 14.02 & — & 83.24 & 0.56 & 22.96 & — \\  \cline{2-10}\cline{2-10}\cline{2-10}
        \multirow{3}{*}{\rotatebox{90}{Learning}}
        & DGR & 77.12 & 1.64 & 33.10 & 4.51 & 96.90 & 0.34 & 21.70 & 73.60 \\ 
        & PointDSC & 97.95 & \textbf{0.30} & 7.23 & 88.03 & 97.96 & \textbf{0.33} & 21.29 & 85.47 \\ 
        & PG-Net & 98.72 & 0.31 & 7.22 & 90.84 & 97.95 & \textbf{0.33} & 21.29 & 83.82 \\
        \cline{2-10}\cline{2-10}\cline{2-10}
        & OURS & \textbf{98.96} & \textbf{0.30} & \textbf{7.17} & \textbf{91.52} & \textbf{98.21} & \textbf{0.33} & \textbf{21.26} & \textbf{84.31} \\ 
        \noalign{\hrule height 6pt} 
    \end{tabular}
    }
    \caption{Quantitative Results on the KITTI dataset.}
    \label{tab:example2}  
\end{table}

To further confirm the effectiveness of GPI-Net in outdoor scenarios, we assess GPI-Net using the KITTI dataset. It is worth noting that our GPI-Net surpasses all these methods across multiple metrics in Table \ref{tab:example2}.
Specifically, by utilizing the FPFH descriptor, our GPI-Net reaches a 24.55\% improvement in RR over RANSAC, and with the FCGF descriptor, the gain is 17.85\%. Meanwhile, our method preserves lower RE and TE while attaining the highest RR when paired with the FPFH descriptor.
Furthermore, our method delivers competitive gains in RR and F1 relative to PointDSC, while achieving a lower TE. In comparison to PG-Net, our method not only attains higher RR and F1 but also minimizes errors with the FPFH descriptor.
In summary, the results from both indoor and outdoor scenarios fully demonstrate the superiority and robustness of our network.

\subsection{Ablation Study}

In this section, we assess the effectiveness of the design decisions behind the block structures of GPI-Net. Table \ref{tab:example3} gives the performance of different block combinations within the GPI-Net with the FPFH descriptor. The first row represents PG-Net, serving as a baseline for comparison. It is evident that all block combinations outperform the baseline.

For instance, incorporating the DMG improves RR by 0.81\% on the 3DMatch dataset, indicating that the DMG effectively aggregates local and global information through an innovative multi-granularity mixing strategy with dual-path parallel interactions.
Next, adding the GFA on top of the DMG further boosts RR by 1\%, highlighting that both GFA and DMG extract rich contextual information. This facilitates the separation of inliers and outliers.
Ultimately, the inclusion of the OI contributes to an overall improvement of approximately 1.5\%. The OI eliminates redundant information through orthogonal integration strategy, strengthening the power to recognize inliers, and significantly boosting the accuracy and robustness of point cloud registration.

\newcommand{\thickhlines}{\noalign{\hrule height 2pt}}
\renewcommand{\arraystretch}{1.1} 
\begin{table}[tb]
    \centering
    \resizebox{\columnwidth}{!}{ 
    \small 
    \begin{tabular}{ccccccc}
    \thickhlines
        \multirow{2}{*}{datasets} & \multirow{2}{*}{baseline} & \multirow{2}{*}{OI} & \multirow{2}{*}{GFA} & \multirow{2}{*}{DMG} & \multicolumn{2}{c}{\textbf{FPFH}} \\ \cline{6-7}
        ~ & ~ & ~ & ~ & ~ & RR(\%↑) & F1(\%↑)  \\ \cline{1-7}
         \multirow{5}{*}{3DMatch} 
        & \checkmark & ~ & ~ & ~ & 81.96 & 74.82 \\
        ~ & ~ & \checkmark & ~ & ~ & 82.13 & 74.94  \\ 
        ~ & ~ & ~ & \checkmark & ~ & 82.18 & 74.89 \\
        ~ & ~ & ~ & ~ & \checkmark & 82.77 & 75.34 \\ 
        ~ & ~ & ~ & \checkmark & \checkmark & 82.99 & 75.46 \\ 
        ~ & ~ & \checkmark & \checkmark & \checkmark & \textbf{83.25} & \textbf{75.57} \\ \cline{2-7}
         \multirow{5}{*}{KITTI} 
        & \checkmark & ~ & ~ & ~ & 98.72 & 90.84 \\
        ~ & ~ & \checkmark & ~ & ~ & 98.72 & 90.84  \\
         ~ & ~ & ~ & \checkmark & ~ & 98.76 & 90.88 \\
        ~ & ~ & ~ & ~ & \checkmark & 98.86 & 91.23 \\ 
        ~ & ~ & ~ & \checkmark & \checkmark & 98.92 & 91.42 \\ 
        ~ & ~ & \checkmark & \checkmark & \checkmark & \textbf{98.96} & \textbf{91.52} \\ \cline{2-7}
    \thickhlines
    \end{tabular}
    }
    \caption{Ablation Study. \textbf{Baseline}: PG-Net. \textbf{OI}: Orthogonal Integration. \textbf{GFA}: Gestalt Feature Attention. \textbf{DMG}: Dual-path Multi-Granularity parallel interaction aggregation.}
    \label{tab:example3}  
\end{table}

\section{Conclusion}

In this paper, we propose an efficient and robust GPI-Net based on Gestalt principles. Specifically, the OI leverages the spatial information of feature vectors and adopts an orthogonal integration strategy to optimally reduce redundant information during feature fusion. The GFA employs a combination of self-attention and cross-attention to precisely obtain the geometric features of correspondences. The DMG performs an innovative multi-granularity feature extraction via dual-path parallel interaction strategy and facilitates information exchange across granularities, effectively grasping the relationship between local details and global information. GPI-Net promotes a more accurate differentiation between inliers and outliers. Extensive experiments demonstrate that our proposed GPI-Net delivers satisfactory improvements in both performance and robustness.

\section*{Acknowledgments}
This work was supported in part by National Natural Science Foundation of China (Grant Nos. 62171130 and 62301160). 

\section*{Contribution Statement}

Weikang Gu and Mingyue Han contributed equally to this work.   
Lifang Wei is the corresponding author.

\bibliographystyle{named}
\bibliography{ijcai25}

\end{document}